# ModelHub.AI:
# Dissemination Platform for Deep Learning Models

Ahmed Hosny[*†,1,2,4], Michael Schwier[*3,4], Christoph Berger[1,5], Evin P Örnek[5], Mehmet Turan[6], Phi V Tran[7], Leon Weninger[8], Fabian Isensee[9], Klaus H Maier-Hein[9], Richard McKinley[10], Michael T Lu[1,4,11,12], Udo Hoffmann[1,4,11,12], Bjoern Menze[5], Spyridon Bakas[13,14], Andriy Fedorov[3,4], Hugo JWL Aerts[‡1,2,3,4,11,15]

[1]Artificial Intelligence in Medicine (AIM) Program, Brigham and Women's Hospital, Harvard Medical School, Boston, MA, USA

[2]Radiation Oncology & [3]Radiology, Brigham and Women's Hospital, Dana-Farber Cancer Institute, Harvard Medical School, Boston, MA, USA

[4]Harvard Medical School, Boston, MA, USA

[5]Institute for Advanced Study, Department of Informatics, Technical University of Munich, Munich, Germany

[6]Max Planck Institute for Intelligent Systems, Stuttgart, Germany

[7]Booz | Allen | Hamilton, McLean, VA, USA

[8]Institute of Imaging & Computer Vision, RWTH Aachen University, Aachen, Germany

[9]Division of Medical Image Computing, German Cancer Research Center (DKFZ), Heidelberg, Germany

[10]Support Centre for Advanced Neuroimaging, University Institute of Diagnostic and Interventional Neuroradiology Inselspital, Bern University Hospital, Bern, Switzerland

[11]Cardiovascular Imaging Research Center, Massachusetts General Hospital, Harvard Medical School, Boston, USA

[12]Cardiac MR PET CT Program, Department of Radiology, Massachusetts General Hospital, Boston, MA, USA

[13]Center for Biomedical Image Computing and Analytics, University of Pennsylvania, Philadelphia, PA, USA

[14]Radiology & Pathology and Laboratory Medicine, Perelman School of Medicine, University of Pennsylvania, Philadelphia, PA, USA

[15]Radiology and Nuclear Medicine, GROW & CARIM, Maastricht University, Maastricht, Netherlands

[*]Authors contributed equally

† ahmed_hosny@dfci.harvard.edu   ‡ haerts@bwh.harvard.edu

*Abstract*— Recent advances in artificial intelligence research have led to a profusion of studies that apply deep learning to problems in image analysis and natural language processing among others. Additionally, the availability of open-source computational frameworks has lowered the barriers to implementing state-of-the-art methods across multiple domains. Albeit leading to major performance breakthroughs in some tasks, effective dissemination of deep learning algorithms remains challenging, inhibiting reproducibility and benchmarking studies, impeding further validation, and ultimately hindering their effectiveness in the cumulative scientific progress. In developing a platform for sharing research outputs, we present ModelHub.AI *(www.modelhub.ai)*, a community-driven container-based software engine and platform for the structured dissemination of deep learning models. For contributors, the engine controls data flow throughout the inference cycle, while the contributor-facing standard template exposes model-specific functions including inference, as well as pre- and post-processing. Python and RESTful Application programming interfaces (APIs) enable users to interact with models hosted on ModelHub.AI and allows both researchers and developers to utilize models out-of-the-box. ModelHub.AI is domain-, data-, and framework-agnostic, catering to different workflows and contributors' preferences.

## I. Introduction

The generation of large amounts of data, availability of specialized computational hardware, as well as advancements in machine learning, have all led to the recent profusion of artificial intelligence (AI) applications in fields ranging from computer vision [1] and natural language processing [2] to radiology [3] and beyond. Moreover, the availability of over a dozen open-source deep learning computational frameworks have immensely lowered the barriers to entry and utilization, in addition to allowing state-of-the-art methods to be implemented in a few lines of code. Despite these reported successes and widespread adoption, issues around the transparency and reproducibility of studies continue to hamper progress in AI research, and have been compared to those that have burdened medicine, psychology, and other fields over the past decade [4]. In a 2018 survey of 400 AI studies, none have reported all the variables needed to reproduce the experiments, only 6% shared code, 30% shared test data, and 54% shared pseudocode [5]. Another effort surveyed 30 text mining studies and reported that important explicit information regarding datasets, study parameters, randomization control, and software environments were lacking in most studies [6]. Finally, and for studies that do share methods, ensuring its longevity and avoiding broken links to online resources is rather challenging: 18% of 704 natural language processing papers that did publish training data referenced links that were broken or deprecated within five years of publication [7].

Shortcomings in the dissemination of AI research outputs are a result of multiple related factors. Unpublished or inadequately published code is perhaps the primary factor and may be a result of lack of resources, the heavy burden of distribution and maintenance, incomplete documentation, as well as intellectual property and licensing restrictions. The highly diverse landscape of frameworks used to perform deep learning studies is another contributing factor, making interoperability and cross-framework implementation rather challenging. Some of these frameworks exhibit fundamentally different computational graph definitions, including declarative "define-and-run" definitions for predefined static graphs, as well as imperative "define-by-run" graphs that are defined dynamically via computation. As a result, transitioning from one to the other comes with a steep learning curve given differences in variable declarations and debugging.

Additionally, the development of deep learning frameworks is also highly volatile given their relatively recent debut. Popular frameworks (e.g., *Torch*, *CNTK*, and *Theano*) are no longer under active major development, while once independent frameworks (e.g., *Caffe2* and *PyTorch*) have now merged into a single library. This volatility introduces uncertainty into the sharing of methods, and ultimately inhibits their utilization and implementation by the community.

Despite these challenges, some authors - together with the wider open-source community - continue to share code and implement studies in an ad-hoc fashion. Although originally designed for software version control, repositories in web-based services (e.g., *GitHub*) host many of these implementations. These repositories are relatively homogenous as they mainly comprise models for object localization and classification in photographic images as part of the ImageNet Large Scale Visual Recognition Challenge[1]. Additionally, given the lack of standardized formats and test cases, together with the wide range of documentation scope and breadth, proper execution of code from these repositories may require significant trial-and-error and ensuring completeness is thus often unattainable. Many deep learning frameworks host collections of working models often referred to as "model zoos". However, these are almost exclusively comprised of models from high profile studies only, are framework-specific by default, and occasionally lack documented pre- and post- processing pipelines where only the pretrained model is shared. As a last resort for studies that lack code implementations, some efforts have turned to automating information extraction from figures, diagrams, and tables in research manuscripts and converting these into abstract computational graphs

---

[1] http://www.image-net.org/challenges/LSVRC/

[8]. Finally, and while significant efforts have been aimed at curating data for public dissemination through machine learning competitions and challenges [9,10], no comparable efforts are being made in the inverse direction of making solutions - that have been developed using this data - publicly available. As such, a systematic 'backward translation' of models into research is much needed.

In this study, we develop a medium for the sharing of deep learning research outputs, in line with the FAIR principles [11] and with a focus on transparency, reproducibility, and ease of both contribution and consumption. As a complement to scientific manuscripts, we present ModelHub.AI (further referred to as ModelHub), a community-driven software engine and template for the structured dissemination of deep learning models. For contributors, the ModelHub engine controls data flow throughout the inference cycle, as well as input loading and type casting out-of-the-box. The engine is also extensible allowing for processing multiple data types, making it both domain- and data-agnostic. On the other hand, the contributor-facing ModelHub template allows for focus on implementation-specific operations including inference and pre- and post-processing, and thereby reducing the effort and time required to share code. On the receiving end, users interact with models hosted on ModelHub through standard *Python* and RESTful application programming interfaces (APIs) that return information about a given model or inference on a given input. This facilitates direct integration into users' benchmarking routines and additional validation studies. ModelHub is also container-based. In addition to the added portability, this allows ModelHub to cater to different frameworks and preferences. ModelHub aims to be a repository of deep learning models with accompanying scientific manuscripts allowing for the pairing of these two distinctly different research dissemination media. Finally, ModelHub is open-source enabling more advanced users to further explore containerized models beyond the API, and giving contributors flexibility in choosing appropriate model sharing licenses.

II. ARCHITECTURE

The ModelHub architecture (Fig. 1) comprises three main components: a containerization scheme for runtime environment control, an engine for data flow and processing, and a model-specific template provided by the respective contributors.

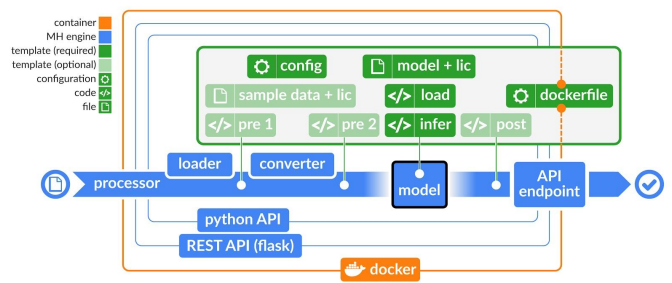

**Fig. 1.** Overall ModelHub architecture illustrating the container (orange), engine (blue), and model template populated by contributors (green).

*A. Containerization and Runtime Environments*

ModelHub uses *Docker*[2] containers (Fig. 1, orange) for executing models, an industry standard for micro-service virtualization. *Docker* and other container infrastructures (e.g., *rkt*[3]) were originally designed for application deployment. ModelHub utilizes a stacked *Docker* image configuration (Fig. 2). Starting with a base operating system image, the *model image* is built to encompass the contributed model's runtime environment. This often includes the deep learning framework, as well as other *Linux* packages and *Python* libraries. The *ModelHub image* encompasses the *model image* and adds the ModelHub engine as well as its runtime environment. The *deployment image* is optional and allows for incorporating the model source for a fully contained image that can be directly deployed into container orchestration systems (e.g., *Kubernetes*[4]). This stacked design features images with gradually increasing volatility: images containing model runtime being less volatile and unlikely to change - and conversely the actively developed ModelHub engine being more volatile. As such, ModelHub engine and API updates can be backward compatible with existing models by simply rebuilding images. Finally, the decoupling of the contributed model runtime environment from its source files (*model image* vs *deployment image* respectively) enables source files to be hosted and updated in an efficient and isolated manner, in addition to allowing the model runtime environment to be reused across multiple different models with identical requirements.

---
[2] https://www.docker.com/
[3] https://coreos.com/rkt/

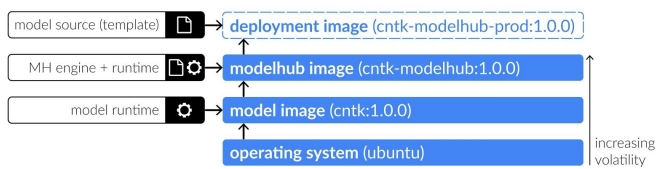

**Fig. 2.** Stacked *Docker* image configuration allows for backwards compatibility and reusable images across models.

### B. ModelHub Engine

The ModelHub engine (Fig. 1, blue) comprises 4 main classes. The *loader* and *converter* classes handle the loading of input data and conversions to *numpy* arrays respectively. Separating these responsibilities allows for pre-processing of data in its native format, in array format, or in both as per the contributor's pipeline. The *Python Imaging Library* (PIL)[5] and *SimpleITK* (a *Python* wrapper for a subset of the *Insight Segmentation and Registration Toolkit* (ITK) functionality) [12] are two currently employed libraries. Future extensions and additional libraries are possible given the chain-of-responsibility pattern employed in the design. The third class is the *model* class where model initialization, loading, and inference functions are housed. Finally, the *processor* class handles data flow throughout the entire inference cycle: loading, converting, pre-processing, and feeding data into the model, in addition to post-processing.

Access to the model and its features is possible through a *Python* API that interfaces with the ModelHub engine. The API handles inference and provides convenience functionalities for access to model configurations and files. A RESTful web API encapsulates the *Python* API and allows for interactions with packaged models through HTTP. As such, it is powered by a *Flask* server running within the container. Table 1 illustrates some examples of the RESTful API endpoints. Both *Python* and RESTful API documentation can be found online[6].

| Endpoint | Method | Returns |
| --- | --- | --- |
| /get_config | GET | model configuration including metadata, manuscript information, model input/output formats and dimensions |
| /get_legal | GET | model and sample data license information |
| /get_model_files | GET | zip folder containing model and associated files |
| /get_samples | GET | urls to sample data |
| /predict_sample | GET | inference result on sample data |
| /predict | GET POST | inference result on input provided through url (GET) or upload (POST), model metadata, processing time |

**Table 1.** Examples of endpoints implemented in the RESTful API. With the exception of /get_model_files, all endpoints return a json response. Information pertaining to sample data is only available if provided by contributor.

In terms of I/O, inputs are managed by the aforementioned *loader* class while ensuring the input adheres to file formats and dimensions predetermined by the contributor. Similarly, outputs are preassigned a type (Fig. 3) and can hence be handled appropriately within other applications interfacing with the ModelHub API. All outputs are returned within json responses. For output types that are not json serializable, a url to an HDF5[7] (.h5) file is returned. The HDF5 format allows for the attachment of named attributes to data (e.g., free-form text description of outputs) and enhances cross-compatibility (e.g., there are HDF5 implementations for .NET[8] and JavaScript[9] applications).

---

[4] https://kubernetes.io/
[5] https://github.com/python-pillow/Pillow
[6] https://modelhub.readthedocs.io/en/latest/modelhubapi.html

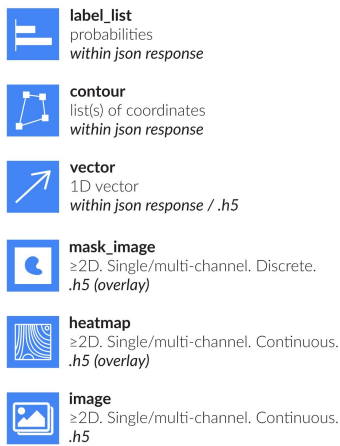

**Fig. 3.** Predefined inference output types enable appropriate handling of data in applications interfacing with the ModelHub API.

*C. Template*

The template (Fig. 1, green) is the contributor-facing component of ModelHub. Templates are to be populated by model source files. At the outset, a *dockerfile* is to be provided as part of the template for building the model runtime environment (Fig. 2, *model image*). The minimum requirements (Fig. 1, dark green) for a valid contribution to the ModelHub model registry comprise the following: a pretrained model file containing both architecture and weights together with license information, *Python* functions for loading and inferring on given inputs, and finally a pre-structured JSON configuration file that contains a unique identifier, model provenance and associated manuscript metadata, as well as I/O format requirements. The configuration file schema can be found online[10]. Other optional components within the template (Fig. 1, light green) include functions for pre- and post-processing of data before and after inference respectively, as well as sample data with associated license information. Documentation for contributing models to the ModelHub registry can be found online[11].

### III. USAGE

The ModelHub infrastructure relies on researchers either commiting new models to the ModelHub registry (contributors) or consuming existing models (users). Given this community-driven aspect, a user-centric approach has been employed in the design of both the model contribution and consumption pipelines.

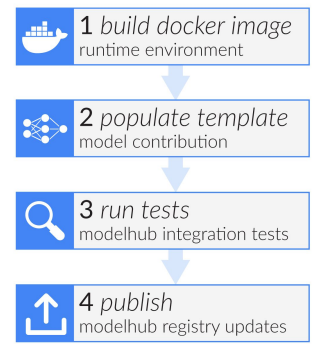

**Fig. 4.** Model contribution pipeline.

*A. For Contributors*

The model contribution pipeline (Fig. 4) commences with the packaging of model source files through creating a *Docker* runtime environment and populating the model template. Model source files are housed in individual repositories. As such, they may be independently versioned, hosted on any web-based version control platform where ownership is preserved, as well as used in a standalone fashion independent of the ModelHub framework or with other model dissemination platforms. The following step involves running ModelHub integration tests that ensure the contribution adheres to specific standards, and API calls return expected results. Finally, contributions are added to the ModelHub registry by reference to the model repository, and made available for consumption after review by the ModelHub core team.

---

[7] https://www.hdfgroup.org/solutions/hdf5/
[8] http://hdf5.net/
[9] https://www.npmjs.com/package/hdf5
[10] https://github.com/modelhub-ai/modelhub/blob/master/config_schema.json
[11] https://modelhub.readthedocs.io/en/latest/contribute.html

*B. For Users*

Users looking to consume models will begin at the ModelHub model registry. The current registry hosts a diverse set of models built in various deep learning frameworks with different I/O formats, performing tasks ranging from ImageNet image classification to facial detection and emotion recognition models, in addition to survival prediction and organ segmentation models in medical images (Supplementary Table 1). A web application[12] showcasing the entire model registry has been developed as part of the ModelHub infrastructure. It allows users to browse and search models, as well as view their metadata and associated manuscripts. Users may also test-drive models directly in the browser by running inference on provided samples or uploaded inputs. Additionally, the web application utilizes *Netron*[13], a web-based model viewer for visualizing computational graphs in multiple formats. This enables comparing and contrasting architectures across models.

Once a model of interest is identified, the single dependency for consuming ModelHub models includes the one-time installation of *Python 2.7 or 3.6*, *Docker*, and the ModelHub *Python* package[14]. Accessed through a command line, the ModelHub *Python* package allows users to select a model from the ModelHub registry and run it locally. Given the model name, the package handles downloading the corresponding *Docker* image (Fig. 2, *ModelHub image*) and running it as a container whilst mounting the respective populated model template. In addition to providing convenience functionalities including the ability to mount local data and modify port mappings, the package offers two modes of interacting with the model. The first mode is through the RESTful API by means of the *Flask* application, and is more suited for out-of-the-box benchmarking and deployment i.e., for users without much interest in the mechanics of the contributed model. The second mode utilizes the *Python* API which can be accessed through a sandboxed *jupyter* notebook provided as part of the model template. This enables more flexibility in modifying or implementing custom pre- or post-processing functions, as well as running batch inference. Alternatively, more advanced users may freely explore the container and its contents through the *Docker's* interactive bash shell. The ModelHub documentation can be found online[15].

IV. CASE STUDIES

To demonstrate the utility of ModelHub in benchmarking models trained on specific prediction tasks, we perform two case studies. In the first case study, we package 9 ImageNet classification models into the ModelHub template and add them to the registry. We benchmark these models (Table 2) on the ImageNet validation set (n=50K), showcasing how the standard *Python* API[16] can be used to compare models built on different deep learning frameworks (Supplementary Table 1) while achieving accuracies comparable to those reported in the original studies. In the second case study, we packaged 3 medical image segmentation models [13–15] submitted to the 2018 International Multimodal Brain Tumor Segmentation (BraTS) challenge [10,16,17]. By benchmarking these models on the BraTS 2018 test set (n=191) (Fig. 5), we showcase the utility of the ModelHub template in soliciting standardized submissions to machine learning competitions, while easing the burden of curation and evaluation on organizers.

---

[12] http://app.modelhub.ai/
[13] https://github.com/lutzroeder/netron
[14] https://pypi.org/project/modelhub-ai/
[15] https://modelhub.readthedocs.io/en/latest/
[16] https://github.com/modelhub-ai/imagenet-benchmark

| Model Name | Implemented Accuracy | | Reported Accuracy | |
|---|---|---|---|---|
| | top-1 | top-5 | top-1 | top-5 |
| xception [18] | 78.1 | 94.1 | 79.0 | 94.5 |
| inception-v3 [19] | 76.7 | 93.3 | 78.8 | 94.4 |
| densenet [20] | 76.6 | 93.4 | 76.2 | 93.2 |
| resnet-50 [21] | 75.0 | 92.3 | 77.2 | 93.3 |
| vgg-19 [22] | 73.7 | 91.5 | 74.5 | 92.0 |
| mobilenet [23] | 70.9 | 89.9 | 72.0 | n/a |
| googlenet [24] | 68.0 | 88.5 | n/a | 93.3 |
| squeezenet [25] | 56.0 | 78.9 | 57.5 | 80.3 |
| alexnet [26] | 55.8 | 79.1 | 57.2 | 80.3 |

**Table 2** Benchmarking results on 9 models from the ImageNet Large Scale Visual Recognition Challenge, and packaged into the ModelHub template.

## V. Discussion

In this study, we present ModelHub, a publishing platform for the structured dissemination of pre-trained deep learning models. Contributions by the research community result in containerized standalone units that can be executed out-of-the-box. With an initial focus on inference, a standard API allows integration into research studies and other applications. As such, ModelHub aims to lower the barriers to consuming published models by eliminating much of the effort involved in implementing them.

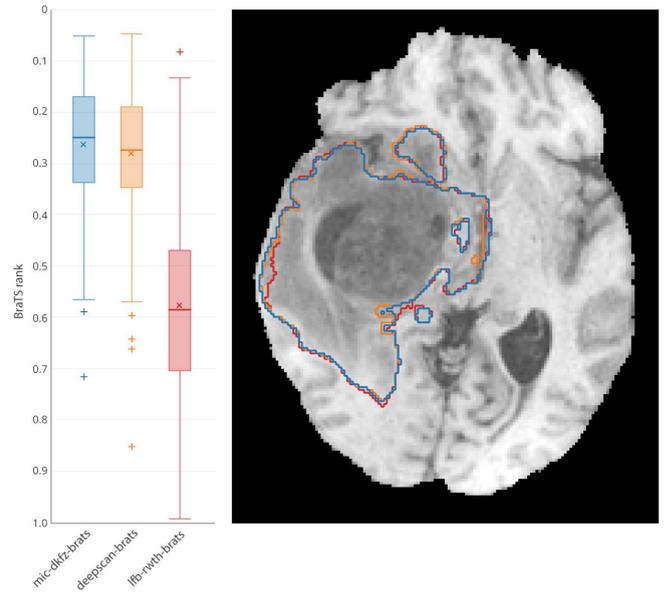

**Fig. 5.** Ranking for 3 brain tumor segmentation models as part of the BraTS challenge - smaller values are higher ranks (left) and resultant contours on a test case (right)

*A. Related Work*

Multiple efforts have proposed solutions to facilitate the sharing of domain-specific deep learning models. Examples of these include Kipoi [27], a repository of machine learning models developed for genomic applications, DeepInfer [28], a deep learning deployment toolkit for image-guided therapy within 3D Slicer [29], the Cancer imaging Phenomics Toolkit (CaPTk) [30], which allows the incorporation of binarized deep learning algorithms for inference, TOMAAT [31], a cloud deployment infrastructure for models trained on volumetric medical images, as well as the deployment capabilities of NVIDIA Clara[17], a machine learning framework for medical imaging workflows. ModelHub is domain-agnostic, promoting transfer learning [30], the inter-domain utilization of containerized pretrained models for many studies with insufficient training data [31]. Other efforts include the *Caffe* model zoo[18],

---
[17] https://docs.nvidia.com/clara/
[18] http://caffe.berkeleyvision.org/model_zoo.html

*Tensorflow* model repository[19] and DLHub [32], as well as the *MXNet* model zoo[20] for the sharing of *Caffe* [33], *Tensorflow/Keras* [34], and *MXNet* [35] models respectively. In addition to framework-specific model formats, contributions to these repositories also feature framework-specific pre- and post-processing pipelines. ModelHub's framework-agnostic design allows researchers to continue utilizing their preferred frameworks. This also bypasses issues regarding the cross-framework interoperability and conversion of models. Modelhub also supports open-source formats aimed at the general representation of deep learning graphs, including the neural network exchange format (NNEF) [36] and the open neural network exchange format (ONNX)[21].

By utilizing containers for packaging models, ModelHub builds upon a large body of work that employs containers in scientific research as a portability and reproducibility mechanism. This trend has motivated the development of specialized container infrastructures including *Singularity* [37], designed to integrate with scientific computational resources such as High Performance Computing (HPC) clusters. Containers facilitate model consumption as familiarity with the underlying deep learning framework and its mechanics is not required. Despite requiring curated contributions within the containers, ModelHub features a lightweight model template that is written in *Python*, simple enough to navigate, populate, and extend.

*B. Limitations*

Several limitations should also be noted. By design, ModelHub only supports inference. Reproducing training pipelines requires knowledge of model hyperparameters, access to training data and associated curation code, controlled model initializations, as well as training runtime environments. In considering trained models as the ultimate research output, the ModelHub template is not designed for experimentation, but rather for packaging, documenting, and distributing models at the conclusion of studies in consistency with the FAIR principles [11]. Nonetheless, we anticipate future releases of ModelHub to include hyperparameter configurations, as well as provisions for sharing training code. Another limitation stems from the dependency on *Docker* that might require extra familiarization effort from novice users, in addition to security concerns as running *Docker* requires users to have root privileges. However, the rising ubiquity of cloud-based applications and software containerization practices in both research and industry, as well as the push towards open industry standards around container formats and runtime[22], both promise widespread adoption.

*C. Improvements*

Future improvements include expanding the model registry in both breadth (variety of domains and data types) and depth (number of contributions), as well as providing support for more input and output data types. Such growth will enable future meta analysis of models where the evolution of model design (e.g., architectures, activation functions, regularizers) over time and across domains can be studied. The contribution process is also to be simplified through continuous integration tests and model registry updates, together with enabling model contribution entirely through a web browser. While ModelHub provides APIs for interacting with models, the serving of said models is self-service where users are required to operate and manage their deployment locally or on cloud infrastructure. Provisions for fully hosted services are in place and may be utilized in future releases, similar to those offered through *Tensorflow serving* [38] and *clipper* [39]. This will involve the integration of user authentication, load-balancing, auto-healing, and other features needed for serving models at scale. As the current versioning of model source files is handled by *Git*, future improvements also include model-specific versioning as proposed by recent deep learning model lifecycle management systems [40]. Finally, we aim to assign a digital object identifier (DOI) [41] to each contribution. As such, models can be identified, cited, and exchanged as intellectual scholarly articles.

---

[19] https://github.com/tensorflow/models
[20] https://mxnet.apache.org/model_zoo/index.html
[21] http://onnx.ai/

[22] https://www.opencontainers.org/

## VI. CONCLUSION

Ensuring accurate and complete dissemination of AI research outputs promises to pave a path towards achieving general AI from the cumulative knowledge gained through the disparate task-specific narrow AI studies conducted today. As many studies remain at the proof-of-concept stage and may lack arguments to demonstrate effectiveness [42], an accurate measure of generalizability and progress in the field may be gauged through the continuous benchmarking of new efforts against existing studies. Nevertheless, one crucial component remains towards effective dissemination: authors being conscientious and willing to invest time and effort in achieving it. Ultimately, it is the unequivocal communication of ideas, code, and data that will bring much needed transparency to computational research experiments, and science en masse.

## ACKNOWLEDGMENTS

Authors acknowledge financial support from the National Institute of Health (NIH:U24CA194354, NIH:U01CA190234, NIH/NCI/ITCR:U01CA242871, NIH/NCI/ITCR:U24CA189523).

## AUTHOR CONTRIBUTIONS
AH, MS, CB, AF, HJWLA - ModelHub core team
EPO, MT - sfm-learner-pose model
PVT - cardiac-fcn model
LW - lfb-rwth-brats model
FI, KHMH - mic-dkfz-brats model
RM - deepscan-brats model
MTL, UH - cxr-prognosis model
BM, SB - BRATS team

## REFERENCES

1. Voulodimos A, Doulamis N, Doulamis A, Protopapadakis E. Deep Learning for Computer Vision: A Brief Review. Comput Intell Neurosci. 2018;2018: 7068349.

2. Young T, Hazarika D, Poria S, Cambria E. Recent Trends in Deep Learning Based Natural Language Processing [Review Article]. IEEE Comput Intell Mag. 2018;13: 55–75.

3. Hosny A, Parmar C, Quackenbush J, Schwartz LH, Aerts HJWL. Artificial intelligence in radiology. Nat Rev Cancer. 2018. doi:10.1038/s41568-018-0016-5

4. Hutson M. Artificial intelligence faces reproducibility crisis. Science. 2018;359: 725–726.

5. Gundersen OE, Kjensmo S. State of the art: Reproducibility in artificial intelligence. Thirty-Second AAAI Conference on Artificial Intelligence. 2018. Available: https://www.aaai.org/ocs/index.php/AAAI/AAAI18/paper/viewPaper/17248

6. Olorisade BK, Brereton P, Andras P. Reproducibility in Machine Learning-Based Studies: An Example of Text Mining. 2017. Available: https://openreview.net/pdf?id=By4l2PbQ-

7. Mieskes M. A quantitative study of data in the nlp community. Proceedings of the First ACL Workshop on Ethics in Natural Language Processing. 2017. pp. 23–29.

8. Sethi A, Sankaran A, Panwar N, Khare S, Mani S. DLPaper2Code: Auto-generation of code from deep learning research papers. Thirty-Second AAAI Conference on Artificial Intelligence. 2018. Available: https://www.aaai.org/ocs/index.php/AAAI/AAAI18/paper/viewPaper/17100

9. Deng J, Dong W, Socher R, Li LJ, Li K, Fei-Fei L. ImageNet: A large-scale hierarchical image database. 2009 IEEE Conference on Computer Vision and Pattern Recognition. 2009. pp. 248–255.

10. Bakas S, Reyes M, Jakab A, Bauer S, Rempfler M, Crimi A, et al. Identifying the Best Machine Learning Algorithms for Brain Tumor Segmentation, Progression Assessment, and Overall Survival Prediction in the BRATS Challenge. arXiv [cs.CV]. 2018. Available: http://arxiv.org/abs/1811.02629

11. Wilkinson MD, Dumontier M, Aalbersberg IJJ, Appleton G, Axton M, Baak A, et al. The FAIR Guiding Principles for scientific data management and stewardship. Sci Data. 2016;3: 160018.

12. Lowekamp BC, Chen DT, Ibáñez L, Blezek D. The Design of

SimpleITK. Front Neuroinform. 2013;7: 45.

13. Weninger L, Rippel O, Koppers S, Merhof D. Segmentation of Brain Tumors and Patient Survival Prediction: Methods for the BraTS 2018 Challenge. Brainlesion: Glioma, Multiple Sclerosis, Stroke and Traumatic Brain Injuries. Springer International Publishing; 2019. pp. 3–12.

14. Isensee F, Kickingereder P, Wick W, Bendszus M, Maier-Hein KH. No New-Net. Brainlesion: Glioma, Multiple Sclerosis, Stroke and Traumatic Brain Injuries. Springer International Publishing; 2019. pp. 234–244.

15. McKinley R, Meier R, Wiest R. Ensembles of Densely-Connected CNNs with Label-Uncertainty for Brain Tumor Segmentation. Brainlesion: Glioma, Multiple Sclerosis, Stroke and Traumatic Brain Injuries. Springer International Publishing; 2019. pp. 456–465.

16. Menze BH, Jakab A, Bauer S, Kalpathy-Cramer J, Farahani K, Kirby J, et al. The Multimodal Brain Tumor Image Segmentation Benchmark (BRATS). IEEE Trans Med Imaging. 2015;34: 1993–2024.

17. Bakas S, Akbari H, Sotiras A, Bilello M, Rozycki M, Kirby JS, et al. Advancing The Cancer Genome Atlas glioma MRI collections with expert segmentation labels and radiomic features. Scientific Data. 2017. doi:10.1038/sdata.2017.117

18. Chollet F. Xception: Deep learning with depthwise separable convolutions. Proceedings of the IEEE conference on computer vision and pattern recognition. 2017. pp. 1251–1258.

19. Szegedy C, Vanhoucke V, Ioffe S, Shlens J, Wojna Z. Rethinking the inception architecture for computer vision. Proceedings of the IEEE Conference on Computer Vision and Pattern Recognition. 2016. pp. 2818–2826.

20. Huang G, Liu Z, Van Der Maaten L, Weinberger KQ. Densely connected convolutional networks. Proceedings of the IEEE conference on computer vision and pattern recognition. 2017. pp. 4700–4708.

21. He K, Zhang X, Ren S, Sun J. Deep residual learning for image recognition. Proceedings of the IEEE conference on computer vision and pattern recognition. 2016. pp. 770–778.

22. Simonyan K, Zisserman A. Very Deep Convolutional Networks for Large-Scale Image Recognition. arXiv [cs.CV]. 2014. Available: http://arxiv.org/abs/1409.1556

23. Sandler M, Howard A, Zhu M, Zhmoginov A, Chen L-C. Mobilenetv2: Inverted residuals and linear bottlenecks. Proceedings of the IEEE Conference on Computer Vision and Pattern Recognition. 2018. pp. 4510–4520.

24. Szegedy C, Liu W, Jia Y, Sermanet P, Reed S, Anguelov D, et al. Going deeper with convolutions. Cvpr; 2015. Available: http://openaccess.thecvf.com/CVPR2015.py

25. Iandola FN, Han S, Moskewicz MW, Ashraf K, Dally WJ, Keutzer K. SqueezeNet: AlexNet-level accuracy with 50x fewer parameters and< 0.5 MB model size. arXiv preprint arXiv:1602 07360. 2016. Available: https://arxiv.org/abs/1602.07360

26. Krizhevsky A, Sutskever I, Hinton GE. ImageNet Classification with Deep Convolutional Neural Networks. In: Pereira F, Burges CJC, Bottou L, Weinberger KQ, editors. Advances in Neural Information Processing Systems 25. Curran Associates, Inc.; 2012. pp. 1097–1105.

27. Avsec Z, Kreuzhuber R, Israeli J, Xu N, Cheng J, Shrikumar A, et al. Kipoi: accelerating the community exchange and reuse of predictive models for genomics. bioRxiv. 2018. p. 375345. doi:10.1101/375345

28. Mehrtash A, Pesteie M, Hetherington J, Behringer PA, Kapur T, Wells WM 3rd, et al. DeepInfer: Open-Source Deep Learning Deployment Toolkit for Image-Guided Therapy. Proc SPIE Int Soc Opt Eng. 2017;10135. doi:10.1117/12.2256011

29. Fedorov A, Beichel R, Kalpathy-Cramer J, Finet J, Fillion-Robin J-C, Pujol S, et al. 3D Slicer as an image computing platform for the Quantitative Imaging Network. Magn Reson Imaging. 2012;30: 1323–1341.

30. Davatzikos C, Rathore S, Bakas S, Pati S, Bergman M, Kalarot R, et al. Cancer imaging phenomics toolkit: quantitative imaging analytics for precision diagnostics and predictive modeling of clinical outcome. J Med Imaging (Bellingham). 2018;5: 011018.

31. Milletari F, Frei J, Aboulatta M, Vivar G, Ahmadi S-A. Cloud deployment of high-resolution medical image analysis with TOMAAT. IEEE J Biomed Health Inform. 2018. doi:10.1109/JBHI.2018.2885214

32. Chard R, Li Z, Chard K, Ward L, Babuji Y, Woodard A, et al. DLHub: Model and Data Serving for Science. arXiv [cs.LG]. 2018. Available: http://arxiv.org/abs/1811.11213

33. Jia Y, Shelhamer E, Donahue J, Karayev S, Long J, Girshick R, et al. Caffe: Convolutional Architecture for Fast Feature Embedding. Proceedings of the 22Nd ACM International Conference on Multimedia. New York, NY, USA: ACM; 2014. pp. 675–678.

34. Abadi M, Agarwal A, Barham P, Brevdo E, Chen Z, Citro C, et al. TensorFlow: Large-Scale Machine Learning on Heterogeneous Distributed Systems. arXiv [cs.DC]. 2016. Available: http://arxiv.org/abs/1603.04467

35. Chen T, Li M, Li Y, Lin M, Wang N, Wang M, et al. MXNet: A Flexible and Efficient Machine Learning Library for Heterogeneous Distributed Systems. arXiv [cs.DC]. 2015. Available: http://arxiv.org/abs/1512.01274

36. Seo B, Shin M, Mo YJ, Kim J. Top-down parsing for Neural Network Exchange Format (NNEF) in TensorFlow-based deep learning computation. 2018 International Conference on Information Networking (ICOIN). 2018. pp. 522–524.

37. Kurtzer GM, Sochat V, Bauer MW. Singularity: Scientific containers for mobility of compute. PLoS One. 2017;12: e0177459.

38. Olston C, Fiedel N, Gorovoy K, Harmsen J, Lao L, Li F, et al. TensorFlow-Serving: Flexible, High-Performance ML Serving. arXiv [cs.DC]. 2017. Available: http://arxiv.org/abs/1712.06139

39. Crankshaw D, Wang X, Zhou G, Franklin MJ, Gonzalez JE, Stoica I. Clipper: A low-latency online prediction serving system. 14th ${USENIX}$ Symposium on Networked Systems Design and Implementation ({NSDI}$ 17). 2017. pp. 613–627.

40. Miao H, Li A, Davis LS, Deshpande A. ModelHub: Towards Unified Data and Lifecycle Management for Deep Learning. arXiv [cs.DB]. 2016. Available: http://arxiv.org/abs/1611.06224

41. Chandrakar R. Digital object identifier system: an overview. The Electronic Library. 2006;24: 445–452.

42. Király FJ, Mateen B, Sonabend R. NIPS - Not Even Wrong? A Systematic Review of Empirically Complete Demonstrations of Algorithmic Effectiveness in the Machine Learning and Artificial Intelligence Literature. arXiv [cs.LG]. 2018. Available: http://arxiv.org/abs/1812.07519

43. Barsoum E, Zhang C, Ferrer CC, Zhang Z. Training deep networks for facial expression recognition with crowd-sourced label distribution. Proceedings of the 18th ACM. 2016. Available: https://dl.acm.org/citation.cfm?id=2993165

44. Tran PV. A fully convolutional neural network for cardiac segmentation

in short-axis MRI. arXiv preprint arXiv:160400494. 2016. Available: http://arxiv.org/abs/1604.00494

45. Christ PF, Elshaer MEA, Ettlinger F, Tatavarty S, Bickel M, Bilic P, et al. Automatic Liver and Lesion Segmentation in CT Using Cascaded Fully Convolutional Neural Networks and 3D Conditional Random Fields. Medical Image Computing and Computer-Assisted Intervention – MICCAI 2016. Springer International Publishing; 2016. pp. 415–423.

46. Hosny A, Parmar C, Coroller TP, Grossmann P, Zeleznik R, Kumar A, et al. Deep learning for lung cancer prognostication: A retrospective multi-cohort radiomics study. PLoS Med. 2018;15: e1002711.

47. Turan M, Ornek EP, Ibrahimli N, Giracoglu C, Almalioglu Y, Yanik MF, et al. Unsupervised Odometry and Depth Learning for Endoscopic Capsule Robots. 2018 IEEE/RSJ International Conference on Intelligent Robots and Systems (IROS). 2018. doi:10.1109/iros.2018.8593623

48. Deng J, Guo J, Xue N, Zafeiriou S. Arcface: Additive angular margin loss for deep face recognition. Proceedings of the IEEE Conference on Computer Vision and Pattern Recognition. 2019. pp. 4690–4699.

49. Wang P, Chen P, Yuan Y, Liu D, Huang Z, Hou X, et al. Understanding Convolution for Semantic Segmentation. 2018 IEEE Winter Conference on Applications of Computer Vision (WACV). 2018. pp. 1451–1460.

50. Redmon J, Farhadi A. Yolov3: An incremental improvement. arXiv preprint arXiv:180402767. 2018. Available: http://arxiv.org/abs/1804.02767

SUPPLEMENTARY MATERIAL

| Model name | Backend | Model format | Task | I/O |
|---|---|---|---|---|
| squeezenet [25] | mxnet | .onnx | ImageNet classification | 2D image / probabilities |
| googlenet [24] | caffe | .caffemodel | | |
| inception-v3 [19] | tensorflow/keras | .h5 | | |
| vgg-19 [22] | mxnet | .onnx | | |
| xception [18] | tensorflow/keras | .h5 | | |
| alexnet [26] | caffe | .caffemodel | | |
| densenet [20] | tensorflow keras | .h5 | | |
| resnet-50 [21] | mxnet | .onnx | | |
| mobilenet [23] | mxnet | .onnx | | |
| emotion-fer-plus [43] | cntk | .onnx | Facial Expression Recognition | 2D image / probabilities |
| cardiac-fcn [44] | tensorflow/keras | .h5 | Segmenting the right ventricle in MRI | DICOM / 2D mask |
| cascaded-fcn-liver [45] | caffe | .caffemodel | Liver and liver lesion segmentation | DICOM / 2D contour |
| deep-prognosis [46] | tensorflow/keras | .h5 | Predict survival of lung cancer patients | 3D array / probabilities |
| sfm-learner-pose [47] | tensorflow | .pb | Unsupervised Pose & Depth Estimation | 2D image / 1D vector |
| arc-face [48] | mxnet | .onnx | Facial Detection & Recognition | 2D image / 1D vector |
| duc-semantic [49] | mxnet | .onnx | Semantic Segmentation | 2D image / 2D segmentation map |
| yolo-v3 [50] | tensorflow/keras | .h5 | Real-Time Object Detection | 2D image / 3D array |
| lfb-rwth-brats [13] | pytorch | .pth | Brain Tumor Segmentation on MRI | multi NIfTI / 3D segmentation map |
| mic-dkfz-brats [14] | pytorch | .model | | |
| deepscan-brats [15] | pytorch | .pth | | |

**Supplementary Table 1.** Models currently included into the ModelHub registry, illustrating the various backends, model formats, tasks, and I/O formats supported by ModelHub.